\documentclass{article}
\usepackage{PRIMEarxiv}
\usepackage[utf8]{inputenc} 
\usepackage[T1]{fontenc}    
\usepackage{hyperref}       
\usepackage{url}            
\usepackage{booktabs}       
\usepackage{amsfonts}       
\usepackage{nicefrac}       
\usepackage{microtype}      
\usepackage{lipsum}
\usepackage{fancyhdr}
\usepackage{amsmath,amssymb,mathtools}
\usepackage{algorithm}
\usepackage{algpseudocode}
\usepackage{graphicx,float}
\usepackage{subfig}
\usepackage{bm} 
\usepackage{amsthm}
\newtheorem{theorem}{Theorem}[section]
\newtheorem{lemma}[theorem]{Lemma}
\usepackage{xcolor}
\usepackage{bm}
\newtheorem{definition}{Definition}
\hypersetup{
    colorlinks,
    linkcolor={red!50!black},
    citecolor={blue!50!black},
    urlcolor={blue!80!black}
}
\algrenewcommand\algorithmicrequire{\textbf{Input:}}
\algrenewcommand\algorithmicensure{\textbf{Output:}}
\DeclareMathOperator*{\argmin}{\arg\!\min}

\title{Improved Conformalized Quantile Regression
\thanks{\textit{\underline{Remark}}: 
\textbf{This is a preprint whose final form is not yet published.}}
}

\author{
  Martim Sousa \\
  IEETA/DETI, University of Aveiro\\
  \texttt{martimsousa@ua.pt} \\
  \AND
   Ana Maria Tomé\\
    IEETA/DETI, University of Aveiro\\
     \texttt{ana@ua.pt} \\
   \AND
   José Moreira\\
    IEETA/DETI, University of Aveiro\\
    \texttt{jose.moreira@ua.pt} \\
}

\begin{document}
\maketitle
\begin{abstract}
Conformalized quantile regression is a procedure that inherits the advantages of conformal prediction and quantile regression. That is, we use quantile regression to estimate the true conditional quantile and then apply a conformal step on a calibration set to ensure marginal coverage. In this way, we get adaptive prediction intervals that account for heteroscedasticity. However, the aforementioned conformal step lacks adaptiveness as described in (Romano et al., 2019). 
To overcome this limitation, instead of applying a single conformal step after estimating conditional quantiles with quantile regression, we propose to cluster the explanatory variables weighted by their permutation importance with an optimized K-means and apply k conformal steps.
To show that this improved version outperforms the classic version of conformalized quantile regression and is more adaptive to heteroscedasticity, we extensively compare the prediction intervals of both in open datasets. 
\end{abstract}

\keywords{Conformal prediction \and Conformalized quantile regression \and Conditional coverage \and Group-balanced conformal prediction \and Permutation importance}

\section{Introduction}

Conformal prediction (CP) is a set of distribution-free and model agnostic algorithms devised to predict with a user-defined confidence with coverage guarantee, see Theorem (\ref{theorem1}). CP is key in high risk decision-making settings where the true output must be within the prediction interval (PI) with high probability. Unlike bayesian models \cite{1,2,3} that only ensure asymptotic coverage guarantees and bootstrap uncertainty estimations methods \cite{4,5}, CP provides valid coverage in finite samples that work for any model, any distribution, any level $\alpha\in (0,1)$ and any \textit{non-conformity score function} s, under the mild assumption of exchangeability (see Definition (\ref{exchdef})). However, despite being true regardless of the \textit{non-conformity score function} s, it has a direct impact in prediction intervals (PIs) width and adaptiveness to heteroscedasticity. Therefore, we must clerverly choose the \textit{non-conformity score function} s to have short as possible adaptive PIs. PIs are said to be adaptive when they take in consideration the uncertainty of the conditional distribution $Y|X$. In plain words, this means that PIs must vary on the input and be wide whenever the model is highly uncertain or the input is indeed hard to predict and narrow if the model has minimal uncertainty on the given input. Unfortunately, conditional coverage, see Eq.(\ref{condcoverageeq}), is a stronger property that CP does not ensure; however, there are heuristic ways to approximate it. 

CP has two main forms: (i) \textit{inductive conformal prediction} (ICP) and (ii) \textit{full conformal prediction} \cite{6}. In a nutshell, the former requires data splitting and therefore is more scalable, whereas the latter does not require data splitting at the cost of refitting the model multiple times, thus being computationally onerous. Henceforth, in the interest of scalability, we will solely dedicate to ICP, also referred to as \textit{split conformal prediction}.

Throughout this paper, we use the notation depicted in Table (\ref{NotationTable}).

\begin{table}[H]
    \centering
\begin{tabular}{ll}
\toprule
Notation & Meaning \\
\midrule
$\alpha$ & Miscoverage rate $\alpha$ \\
$\hat{C}_{1-\alpha} (\bm{x_{n}})$ & $1-\alpha$ prediction interval on input $\bm{x_n}$ \\
$x_{(k)}$ & \textit{k}-th rank statistic of $(x_1,...,x_n)$\\
$\bm{\epsilon}$ & set of \textit{non-conformity scores} \\
Quantile($\bm{\epsilon}$; $1-\alpha$) & $1-\alpha$ order quantile of $\bm{\epsilon}$  \\
$x$ &  a scalar \\
$\bm{x}$ &  a vector \\
$X$ &  a matrix or a random variable (easily distinguished) \\
$\bm{X}$ & a tensor or a random vector (easily distinguished)\\
$(\bm{x_{train}},\bm{x_{cal}},\bm{x_{val}})$ & train, calibration and test data, respectively\\
\bottomrule
\end{tabular}
\caption{Notation used throughout this paper.}
\label{NotationTable}
\end{table}

\subsection*{Paper outline}

The remainder of this Section does a brief recap on ICP and quantile regression (QR) since the contribution is grounded on it. Section \ref{sec2} summarizes conformalized quantile regression (CQR) as described in (Romano et al., 2019) \cite{11}. Section \ref{sec3} introduces our improved version of CQR for tabular data named improved conformalized quantile regression (ICQR). Section \ref{sec4} extensively compare PIs width and coverage of CQR and ICQR on open datasets and discusses the results. Finally, Section \ref{sec5} draws the main points and findings of this research, and addresses shortcomings.

\subsection{Inductive conformal prediction}

The general outline of ICP is as follows:

\begin{enumerate}
    \item Train a model $\hat{f}$ on a training dataset or use a pre-trained off-the-shelf model;
    \item Define a symmetric heuristic notion of uncertainty denoted as 
    $s: \mathcal{X} \times \mathcal{Y} \rightarrow \mathcal{A} \subseteq \mathbb{R}$ usually referred to as the \textit{non-conformity score function}, where larger scores encode worse agreement between pairs $(\bm{x},y)\in \mathcal{X} \times \mathcal{Y}$;
    \item Compute $\epsilon_1,...,\epsilon_n=s(\bm{x_{1}},y_1),...,s(\bm{x_{n}},y_n)$ \textit{non-conformity scores} on a calibration dataset, not seen by the model during training, using the trained model $\hat{f}$, and the \textit{non-conformity score function} s applied on pairs of a calibration dataset $ \mathcal{D}_{cal}:=\{(\bm{x_i},y_i)\}_{i=1}^{n}$;
    \item Compute $\hat{q}= \text{Quantile} \left( \bm{\epsilon};  \frac{\lceil (n+1)(1-\alpha)\rceil}{n}\right)$;
    \item Deploy PIs as $\hat{C}_{1-\alpha}(\bm{x_{n+1}})=\{y \in \mathcal{Y}: s(\bm{x_{n+1}},y) \leq \hat{q}\}$.
\end{enumerate}

\begin{definition}[Exchangeability]
\label{exchdef}
A sequence of random variables $Z_1,Z_2,...,Z_n \in \mathcal{Z}$ are exchangeable if and only if for any permutation $\pi: \{1,2,...,n\} \rightarrow \{1,2,...,n\}$ and every measurable set $E \subseteq \mathcal{Z}^n$, we have
\begin{equation}
    \mathbb{P}\{(Z_1,Z_2,...,Z_n) \in E\}= \mathbb{P}\{(Z_{\pi(1)},Z_{\pi(2)},...,Z_{\pi(n)}) \in E\}
\end{equation}
\end{definition}

\begin{lemma}
\label{lemmabeforetheorem}
Let $Z_1,...,Z_n \in \mathcal{Z}$ be exchangeable random variables with no ties almost surely, then their ranks are uniformly distributed on \{1,...,n\}.
\begin{proof}
    Let $R_i=\text{Rank}(Z_i)$. Since $(Z_1,...,Z_n)$ are exchangeable, we have a total of n! possible permutations between them all being equally probable. Additionally, given that $Z_1,...,Z_n$ have no ties almost surely, then $\mathbb{P}(R_i\neq R_j)=1, \quad \forall i \neq j \in \{1,...,n\}$. For any $i\in\{1,...,n\}$, if we fix $R_i$ on rank $j\in \{1,...,n\}$, there are (n-1)! possible permutations among the other n-1 random variables. 
    Hence, $\mathbb{P}(R_i=j)=\frac{(n-1)!}{n!}=\frac{1}{n}$ and thus
\begin{equation*}
    \mathbb{P}(R_i=1)=\mathbb{P}(R_i=2)=...=\mathbb{P}(R_i=n)=\frac{1}{n}, \quad \forall i \in\{1,...,n\}.
\end{equation*}
\end{proof}
\end{lemma}

\begin{theorem}[Marginal coverage guarantee]
\label{theorem1}
Let $(\bm{X_1},Y_1),...,(\bm{X_{n}},Y_{n})$ be exchangeable random vectors with no ties almost surely drawn from a distribution P, additionally if for a new pair $(\bm{X_{n+1}},Y_{n+1})$, $(\bm{X_1},Y_1),...,(\bm{X_{n+1}},Y_{n+1})$ are still exchangeable, then by constructing $C_{1-\alpha}(\bm{X_{n+1}})$ using ICP, the following inequality holds for any \textit{non-conformity score function} $s:\mathcal{X} \times \mathcal{Y} \rightarrow \mathcal{A} \subseteq \mathbb{R}$ and any $\alpha \in (\frac{1}{n+1},1)$
\begin{equation}
    1-\alpha\leq \mathbb{P}\{Y_{n+1} \in C_{1-\alpha}(\bm{X_{n+1}})\} \le 1-\alpha+\frac{1}{n+1}.
\end{equation}

\begin{proof}
Since $\bm{Z_1}=(\bm{X_1},Y_1),...,\bm{Z_n}=(\bm{X_n},Y_n)$ are exchangeable random vectors with no ties almost surely, the corresponding \textit{non-conformity scores} $\bm{\epsilon}= \{(\epsilon_i:=s(\bm{Z_i}))\}_{i=1}^{n} \subseteq \mathcal{A} $ are also exchangeable (see Theorem 3 of \cite{65}) with no ties almost surely.
Since  the non-conformity scores $\epsilon_{1},...,\epsilon_{n}, \epsilon_{n+1}$ are exchangeable, their ranks are uniformly distributed by Lemma (\ref{lemmabeforetheorem}). Therefore, $\text{Rank}(\epsilon_{n+1}) \sim \mathcal{U}\{1,n+1\}$. That is,

\begin{equation*}
    \mathbb{P}\{\epsilon_{n+1} \leq \epsilon_{(k)} \}=  \mathbb{P}\{\text{Rank}(\epsilon_{n+1}) \leq k \}=\frac{k}{n+1}, \quad k \in\{1,...,n+1\}.
\end{equation*}
By definition, since $\hat{q}=\text{Quantile}(\bm{\epsilon} ; \frac{\lceil (n+1)(1-\alpha)\rceil}{n})=\epsilon_{(\lceil(n+1)(1-\alpha) \rceil)}$, follows that
\begin{align*}
    &\mathbb{P}\{Y_{n+1} \in C_{1-\alpha}(\bm{X_{n+1})}\}=
    \mathbb{P}\{y \in \mathcal{Y}: s(\bm{X_{n+1}},y) \leq \hat{q}\}=\mathbb{P}\{\epsilon_{n+1}\leq \epsilon_{(\lceil(n+1)(1-\alpha) \rceil)}\}=\frac{\lceil(n+1)(1-\alpha) \rceil}{n+1}.
\end{align*}
It is easy to note that $1-\alpha \leq \frac{\lceil(n+1)(1-\alpha) \rceil}{n+1} \leq \frac{(n+1)(1-\alpha)+1}{n+1} = 1-\alpha+\frac{1}{n+1}$.
\end{proof}
\end{theorem}
As a consequence of Theorem (\ref{theorem1}), ICP provides asymptotic exact coverage since $\lim_{n \rightarrow +\infty}(1-\alpha+\frac{1}{n+1})=1-\alpha \implies \mathbb{P}\{Y_{n+1} \in C_{1-\alpha}(\bm{X_{n+1}})\}=1-\alpha$.
In practice, $\text{Quantile}(\bm{\epsilon};\frac{\lceil (n+1)(1-\alpha)\rceil}{n})$, is essentially a very minor finite sample correction on the $1-\alpha$ quantile. However, henceforth, for simplicity, we will use $\text{Quantile}(\bm{\epsilon};1-\alpha)$, but bear in mind that $\text{Quantile}(\bm{\epsilon};\frac{\lceil (n+1)(1-\alpha)\rceil}{n})$ is the formally correct form.

\begin{definition}[Conditional coverage]
An ICP procedure guarantees conditional coverage if
\begin{equation}
    \label{condcoverageeq}
        \mathbb{P}(Y_{val} \in C_{1-\alpha}(\bm{X_{val}})|\bm{X}=\bm{x_{val}}) \ge 1-\alpha.
\end{equation}
\end{definition}
\subsection{Naive method}
\label{naivemethod}
The most basic and widely used \textit{non-conformity score function} for regression problems is the absolute error given by
\begin{equation}
    s(\bm{x},y)=|y-f_{\bm{\theta}}(\bm{x})|,
\end{equation}
where $f_{\bm{\theta}}(\bm{x})$ is the model's forecast with respect to input $\bm{x}$. After calculating $\hat{q}$ on the calibration set, PIs come as 
\begin{equation}
    \hat{C}_{1-\alpha}(\bm{x_{val}})=[f_{\bm{\theta}}(\bm{x_{val}})-\hat{q},f_{\bm{\theta}}(\bm{x_{val}})+\hat{q}].
\end{equation}

\begin{figure}[H]
  \begin{minipage}[b]{0.5\linewidth}
    \centering
    \includegraphics[width=0.8\linewidth]{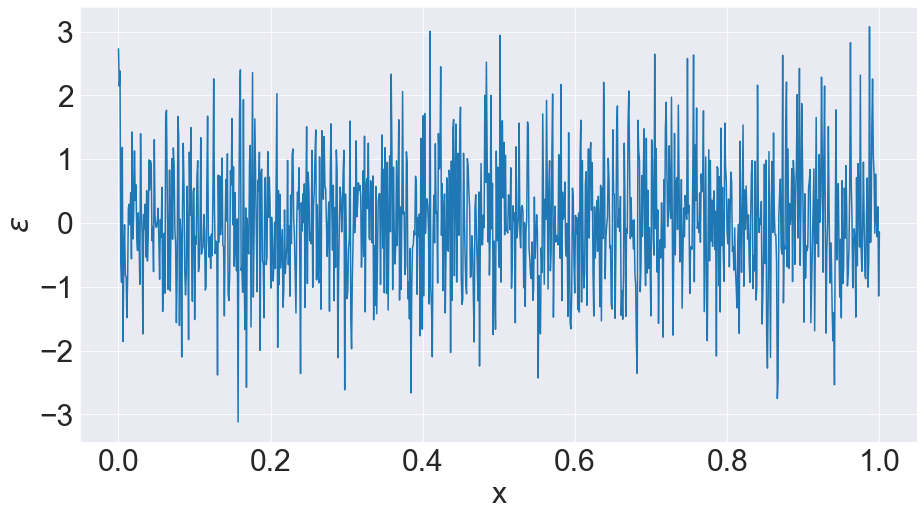} 
    \caption{Homocedastic residuals.}  
    \vspace{4ex}
    \label{homo}
  \end{minipage}
  \begin{minipage}[b]{0.5\linewidth}
    \centering
    \includegraphics[width=0.8\linewidth]{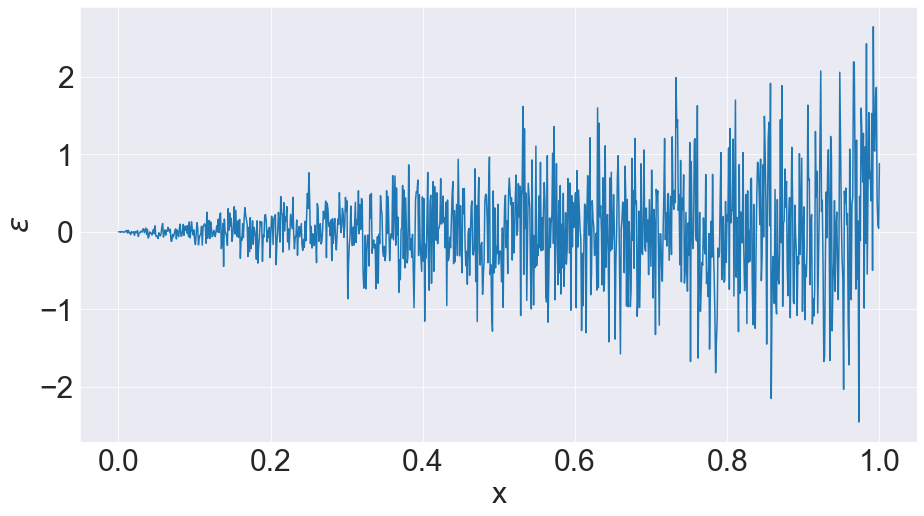} 
    \caption{Heteroscedastic residuals.} 
    \vspace{4ex}
    \label{hetero}
\end{minipage}
\end{figure}

As proven above, this naive method guarantees marginal coverage. However, PIs length are always equal to $2\hat{q}$ regardless of the input, hence not adaptive. In fact, this naive method is known to overcover easy inputs and to undercover hard inputs. As an example, we call the reader's attention to Figs. (\ref{homo}) and (\ref{hetero}). It is straightforward to see that for $\alpha=0.1$ this method would cover the lowest 90\% residuals in Figure (\ref{hetero}) and fail the greatest 10\%. This has great implications in practice since the greatest 10\% \textit{non-conformity scores} might represent a group that is being ignored. In turn, evaluating PIs by simply looking to their mean amplitude or marginal coverage is not enough nor adequate. A finer analysis must look to other statistics as standard deviation and quantiles. PIs with high standard deviation are not necessarily a bad sign, it may sign adaptiveness to heteroscedasticity.

\subsection{Quantile regression}
Usually, in a regression problem, we attempt to find the parameters $\bm{\theta}$ of a model $f_{\bm{\theta}}:\mathbb{R}^{D}\rightarrow \mathbb{R}$ via the minimization of the sum of squared residuals on the training set $\mathcal{D}_{train}$ as
\begin{equation}
    \min_{\bm{\theta}} \sum_{i=1}^{n_{train}} (f_{\bm{\theta}}(\bm{x_i})-y_i)^2 +\Omega(\bm{\theta}),
\end{equation}
where $\Omega(\bm{\theta})$ is a potential regularizer. 

\begin{figure}[H]
    \centering
    \includegraphics[scale=0.3]{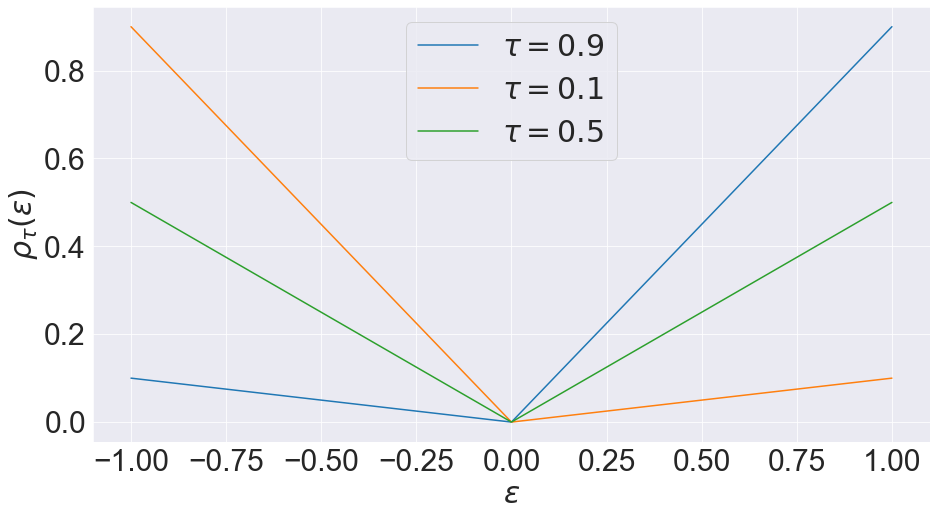}
    \caption{Visualization of the \textit{pinball loss} where $\epsilon=y-f_{\bm{\theta}}(\bm{x})$ for different values of $\alpha$.}
    \label{pinballloss}
\end{figure}

QR \cite{10,12}, however, is a non-parametric method that attempts to approximate the true $\tau \in (0,1)$ conditional quantile $Q_{\tau}(\bm{x})$ given by

\begin{equation}
    Q_{\tau}(\bm{x})=\inf{\{y \in \mathcal{Y}:F_{Y|\bm{X}}(y|\bm{X}=\bm{x}) \ge \tau}\},
\end{equation} of the true conditional distribution $Y|\bm{X}$ by minimizing the following objective
\begin{equation}
    \min_{\bm{\theta}} \sum_{i=1}^{n_{train}} \rho_{\tau}(y_i,f_{\bm{\theta}}(\bm{x_i})) +\Omega(\bm{\theta}),
\end{equation}
where $\rho_{\alpha}$ is the quantile loss, also known as \textit{pinball loss} due to its resemblance to a pinball ball movement, see Fig.(\ref{pinballloss}). This loss function can be mathematically expressed as
\begin{equation}
    \rho_{\tau}(y,f_{\bm{\theta}}(\bm{x}))=\max\left( \tau(y-f_{\bm{\theta}}(\bm{x})), (\tau-1)(y-f_{\bm{\theta}}(\bm{x})\right).
\end{equation}
Consequently, it is suggested to train a QR model $f_{\bm{\theta}}(\bm{x})$ for $\tau=\alpha/2$ and $\tau=1-\alpha/2$ on $\mathcal{D}_{train}$ to obtain $1-\alpha$ conditional coverage via
\begin{equation}
    [\hat{Q}_{\frac{\alpha}{2}}(\bm{x_{val}}),\hat{Q}_{1-\frac{\alpha}{2}}(\bm{x_{val})}],
\end{equation}

where $\hat{Q}_{\frac{\alpha}{2}}(\bm{x_{val}})$ and $\hat{Q}_{1-\frac{\alpha}{2}}(\bm{x_{val}})$ are conditional quantile estimations of $Q_{\frac{\alpha}{2}}(\bm{x_{val}})$ and $Q_{1-\frac{\alpha}{2}}(\bm{x_{val}})$, respectively.  Note that, unlike naive PIs, QR PIs are adaptive (they depend on the input). Another great virtue of QR is that it can be applied on top of any model by just changing the loss function to a \textit{pinball loss}.    
Although the estimation $\hat{Q}_{\alpha}(\bm{x})$ yielded by QR of the unknown true conditional quantile $Q_{\alpha}(\bm{x})$ is known to be asymptotically consistent under certain conditions \cite{11,13,14}, it rarely provides $1-\alpha$ coverage in finite samples. To overcome this limitation, (Romano et al., 2019) \cite{11} drawn several ideas from CP and devised the so-called CQR that we introduce in the next section.

\section{Conformalized quantile regression}
\label{sec2}
CQR grounds on correcting QR intervals with ICP techniques on a calibration set $\mathcal{D}_{cal}$ to ensure marginal coverage, hence inheriting the advantages of both, i.e., adaptive intervals with marginal coverage guarantee. Specifically, if QR bounds are constantly undercovering, then PIs must get wider. On the contrary, in case of QR PIs cover in a ratio superior to $1-\alpha$, they must be shortened. For this purpose, (Romano et al., 2019) \cite{11} proposed the following \textit{non-conformity score function}  
\begin{equation}
\label{q_yhatCQR}
    s(\bm{x},y)=\max{\{\hat{Q}_{\frac{\alpha}{2}}(\bm{x})-y,y-\hat{Q}_{1-\frac{\alpha}{2}}(\bm{x})\}}.
\end{equation}

Subsequently, after calculating $\hat{q}=\text{Quantile}(\bm{\epsilon};1-\alpha)$, adaptive PIs with $1-\alpha$ marginal coverage guarantee are yielded as
\begin{equation}
    [\hat{Q}_{\frac{\alpha}{2}}(\bm{x_{val}})-\hat{q},\hat{Q}_{1-\frac{\alpha}{2}}(\bm{x_{val}})+\hat{q}]
\end{equation}

Built on the same idea, a different \textit{non-conformity score function} was proposed in \cite{21}; however, it has not proved to outperform the \textit{non-conformity score function} in Eq.(\ref{q_yhatCQR}) \cite{22}.
Note that, a $\hat{q}>0$ (most cases) is a result of a QR model that did not ensure $1-\alpha$ coverage and therefore PIs must get wider. In the case of $\hat{q}<0$, it signifies that QR bounds are overcovering and thus we shoul narrow the bounds, i.e., the lower bound increases $\hat{q}$ units and the upper bound decreases the same amount, while still ensuring $1-\alpha$ marginal coverage.

CQR seems appealing, and it is in fact, yet the calibration step lacks adaptiveness. The role of $\hat{q}$ is to simply shift the bounds $\hat{q}$ units. Also, it does not depend on the input at hand in any form. Our contribution, presented in the next section, will focus on improving the calibration step to make it more adaptive and dependent on the input. However, this improvement is only possible in tabular data.

\section{Contribution}
\label{sec3}
\begin{figure}[H]
    \centering
    \includegraphics[scale=0.4]{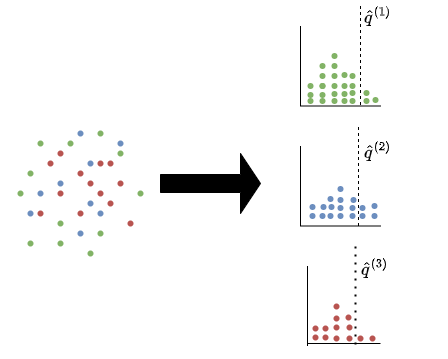}
    \caption{Group-balanced conformal prediction. Image inspired by \cite{8}.}
    \label{groupbalanced}
\end{figure}

Our idea to improve CQR is to do k conformal steps instead, one per each group, as illustrated in Fig.(\ref{groupbalanced}). This is based on the idea that $\bm{x_1} \approx \bm{x_2} \implies f_{\bm{\theta}}(\bm{x_1}) \approx f_{\bm{\theta}}(\bm{x_2})$. To attain such goal, we could compute the euclidean distance between observations and cluster them using K-means \cite{19}; however, this is not a good heuristic to approximate the conditional distribution $Y|X$. We need two additional conditions: (1) features must be scaled, since euclidean distance is scale-dependent; and (2) each feature has a different predictive power upon the response variable y.
To accommodate condition (2), whenever clustering the observations, we must weigh each feature by the respective feature importance. A simple way of calculating feature importance that work for  any model is permutation importance \cite{15}. Algorithms (\ref{kmeansclustalg}), (\ref{PermutationImportanceAlg}), and (\ref{ICQRalg}) comprise every step to successfully perform K-means, permutation importance, and our improved CQR version, respectively. In a nutshell, these are the main steps of our improved version: (1) normalize the training data, and apply the same normalization object on the calibration and validation set; (2) train $f_{\bm{\theta}} :\mathbb{R}^{D} \rightarrow \mathbb{R}$ with \textit{pinball loss} for $\tau=\frac{\alpha}{2}$ and $\tau=1-\frac{\alpha}{2}$; (3) get feature importance by means of permutation importance algorithm using the calibration set for evaluation; (4) create a copy of the training set, calibration set, and validation set weighted by feature importance, henceforth referred to as clustering training set, clustering calibration set, and clustering validation set, respectively; (5) select the best k of K-means on the clustering calibration set and store k centroids to represent each cluster; (6) assign each observation of the clustering calibration set to the nearest cluster/centroid; (7) knowing which elements belong to each of the k clusters given by the previous step, compute a different $\hat{q}$ for each cluster using Eq.(\ref{q_yhatCQR}) as the \textit{non-conformity score function} on the calibration set; (8) given a new observation $\bm{x_{val}}$, find the nearest centroid $\bm{c_i}$ of $\bm{x_{val}^{*}}$ (the matching observation of $x_{val}$ weighted by the respective permutation importance.)  and use the respective $\hat{q}^{(i)}$ to produce PIs as $[\hat{Q}_{\frac{\alpha}{2}}(\bm{x_{val}})-\hat{q}^{(i)},\hat{Q}_{1-\frac{\alpha}{2}}(\bm{x_{val}})+\hat{q}^{(i)}]$.

K-means, as shown in Algorithm (\ref{kmeansclustalg}) is a clustering algorithm that attempts to minimize the following objective
\begin{equation}
\label{kmeansobj}
    L=\sum_{i=1}^{k}\sum_{\bm{x} \in S_i}||\bm{x}-\bm{c_i}||^2,
\end{equation}
where k is the number of clusters; $S_i$ denotes the cluster set i; $c_i$ the centroid of cluster i; and $||.||$ the euclidean norm.

Since K-means relies upon the hyperparameter k to define the number of clusters beforehand, we must select a criteria to find the potential best k. For the purpose at hand, the fraction of variance explained by the centroids is a good choice, which is mathematically expressed as
\begin{equation}
    \label{fracexplainedclust}
    \sigma_k=\frac{\sum_{i=1}^k n_i(\bm{c_i}-\bm{\mu})^T(\bm{c_i}-\bm{\mu})}{\sum_{i=1}^N (\bm{x_i}-\bm{\mu})^T(\bm{x_i}-\bm{\mu})},
\end{equation}
where $n_i$ is the number of examples in cluster i and $\bm{\mu}$ denotes the feature mean.

\subsection*{Remarks}
\begin{itemize}
    \item The selection of k for K-means is not limited to our approach, different criteria regarding the selection of k for K-means are acceptable, e.g., silhouete score  and elbow method \cite{23,24,16};
    \item Different clustering algorithms beyond K-means might also be adequate as long as they take in consideration the essential point, which is cluster based on similarity between observations and feature importance.
\end{itemize}

\begin{algorithm}[H]
\caption{K-means algorithm}
\label{kmeansclustalg}
\begin{algorithmic}[1]
\Require $\bm{c_1}^{(1)},\bm{c_2}^{(1)},...,\bm{c_k}^{(1)}$ initial centroids efficiently randomly assigned as in \cite{20}. 
\Ensure k clusters given by their centroids that minimize Eq.(\ref{kmeansobj})
\State Converged $\gets$ False
\State $t \gets 1$
\While{Converged is False}
\State Assign each observation to the cluster with the nearest centroid i given by
\State $S_i^{(t)}=\{\bm{x}:||\bm{x}-\bm{c_i}^{(t)}|| \le ||\bm{x}-\bm{c_j}^{(t)}|| \; \forall 1\le j \le k\}$
\State $t \gets t+1$
\State Update centroids
\For{$i \gets 1$ to k}
\State $\bm{c_i}^{(t+1)} \gets \frac{1}{|S_i^{(t)}|}\sum_{\bm{x}\in S_i^{(t)}}\bm{x} $
\EndFor
\State Stop if all clusters are the same as from previous iteration.
\If{$\bm{c_i}^{(t+1)}=\bm{c_i}^{(t)}, \quad \forall i \in\{1,2,...,k\}$}
\State Converged $\gets$ True
\EndIf
\EndWhile
\end{algorithmic}
\end{algorithm}

\begin{algorithm}[H]
\caption{Permutation importance}
\label{PermutationImportanceAlg}
\begin{algorithmic}[1]
\Require A dataset $\mathcal{D}$=$\{(\bm{x_i},y_i)\}_{i=1}^N$, a model $f_{\bm{\theta}}: \mathbb{R}^{D} \rightarrow \mathbb{R}$, a performance measure M, and R repetitions. 
\Ensure D feature importance's $I_1,I_2,...,I_D$
\State Split $\mathcal{D}$ in two mutually exclusive sets, a training set $\mathcal{D}_{train}=(X_{train},\bm{y_{train}})$ and a validation set $\mathcal{D}_{val}=(X_{val},\bm{y_{val}})$

\State Train $f_{\bm{\theta}}$ on $\mathcal{D}_{train}$, and compute the baseline error score $E_{b}$ on $\mathcal{D}_{val}$ using the performance measure M
\For{$j \gets 1$ to D}
\For{$i \gets 1$ to R}
\State Permutate column j on the validation X set $X_{val}$ and compute the permutated error score $E_{\pi_j}^{(i)}$ using performance measure M
\State $\Delta_j^{(i)} \gets|E_b-E_{\pi_j}^{(i)}|$
\EndFor
\State $I_j \gets \frac{1}{R}\sum_{i=1}^R \Delta_j^{(i)}$
\EndFor
\end{algorithmic}
\end{algorithm}

\begin{algorithm}[H]
\caption{Improved conformalized quantile regression (ICQR)}
\label{ICQRalg}
\begin{algorithmic}[1]
\Require Three \textbf{normalized} sets: $\mathcal{D}_{train}=(X_{train},\bm{y_{train}}), \mathcal{D}_{cal}=(X_{cal},\bm{y_{cal}})$, and $\mathcal{D}_{val}=(X_{val},\bm{y_{val}})$, \textit{miscoverage error rate} $\alpha$, desired explained variance $\sigma$, maximum number of clusters K, R repetitions, and a model $f_{\bm{\theta}}: \mathbb{R}^{D} \rightarrow \mathbb{R}$
\Ensure Adaptive intervals with $1-\alpha$ coverage

\State Part 1: Estimate conditional quantiles with quantile regression.
\\\hrulefill
\State Use QR to estimate $\hat{Q}_{\frac{\alpha}{2}}(\bm{x})$ and $\hat{Q}_{1-\frac{\alpha}{2}}(\bm{x})$ on $\mathcal{D}_{train}$.
\\\hrulefill
\State Part 2: Weigh the X sets by the feature importance of each feature $I_1,I_2,...,I_{D}$.
\\\hrulefill
\State Get a list of feature importances $I_1,I_2,...,I_D$ given by Algorithm (\ref{PermutationImportanceAlg}) using $\mathcal{D}_{cal}$ for evaluation.
\State $X_{cal}^{*},X_{val}^{*} \gets 0_{n_{cal} \times D},0_{n_{val} \times D}$
\For{$i \gets 1$ to D}
\State col$_i(X_{cal}^{*}) \gets$col$_i(X_{cal}) \times I_i$
\State col$_i(X_{val}^{*}) \gets $col$_i(X_{val}) \times I_i$
\EndFor
\\\hrulefill
\State Part 3: Selection of k for K-means.
\\ \hrulefill
\State $k \gets 2$
\State $best_k \gets$ False
\While{$k\le K$ and $best_k$ is False}
\State Apply K-means as in Algorithm (\ref{kmeansclustalg}) on $X_{cal}^{*}$. 
\State Compute $\sigma_k$ as in Eq.(\ref{fracexplainedclust}).
\If{$\sigma_k>\sigma$}
\State $best_k \gets$ True
\State Store the cluster centroids $\bm{c_1},\bm{c_2},...,\bm{c_k}$
\Else
\State $k \gets k +1$
\EndIf
\EndWhile
\\ \hrulefill
\State Part 4: Compute $\hat{q}^{(1)},\hat{q}^{(2)},...,\hat{q}^{(k)}$ on the calibration set, one per cluster.
\\ \hrulefill
\State $\bm{\epsilon}^{(1)},\bm{\epsilon}^{(2)},...,\bm{\epsilon}^{(k)} \gets \{\},\{\},....,\{\}$
\For{$(\bm{x^{*}},\bm{x},y)$ in $(X_{cal}^{*},X_{cal},\bm{y_{cal}})$}
\State $i \gets \argmin_{j \in \{1,2,...,k\}}{||\bm{x^{*}}-\bm{c_j}||}$
\State $\bm{\epsilon}^{(i)} \gets \bm{\epsilon}^{(i)} \cup \left\{\max\left(\hat{Q}_{\frac{\alpha}{2}}(\bm{x})-y,y-\hat{Q}_{1-\frac{\alpha}{2}}(\bm{x})\right)\right\}$ 
\EndFor
\State $\hat{q}^{(i)} \gets \text{Quantile}\left(\bm{\epsilon}^{(i)};1-\alpha \right), \quad \forall i \in\{1,...,k\}$
\\ \hrulefill
\State Part 5: Deploy PIs on new data. Exemplified as $(X_{val},\bm{y_{val}})$.
\\ \hrulefill
\For{$(\bm{x^{*}},\bm{x},y)$ in $(X_{val}^{*},X_{val},\bm{y_{val}})$}
\State $i \gets \argmin_{j \in \{1,2,...,k\}}{||\bm{x^{*}}-\bm{c_j}||}$
\State  Return $\hat{C}_{1-\alpha}(\bm{x}) \gets [\hat{Q}_{\frac{\alpha}{2}}(\bm{x})-\hat{q}^{(i)},\hat{Q}_{1-\frac{\alpha}{2}}(\bm{x})+\hat{q}^{(i)}]$
\EndFor
\end{algorithmic}
\end{algorithm}

\section{Experiments}
\label{sec4}
In this section we apply Algorithm (\ref{ICQRalg}) and compare it against CQR, QR and Naive on the datasets shown in Table (\ref{datasets}). We use a FFNN (feedforward neural network) \cite{28} with two output neurons to estimate $Q_\frac{\alpha}{2}(\bm{x})$, and $Q_{1-\frac{\alpha}{2}}(\bm{x})$, respectively. This is easily achieved by having a \textit{pinball loss} function with $\tau=\alpha/2$ for the first output neuron and $\tau=1-\alpha/2$ for the second.
$\hat{Q}_\frac{\alpha}{2}(\bm{x})$ and $\hat{Q}_{1-\frac{\alpha}{2}}(\bm{x})$ represent the estimated lower bound, upper bound, with respect to $\bm{x}$, respectively. Due to the stochastic behaviour of FFNN, the model is trained T=100 times to reduce the associated variance of the random initialization. On top of these 100 trained models each of the aforementioned methods is applied. Each dataset in Table (\ref{datasets}) is divided in three mutually exclusive datasets, a training set containing 50\% of the data, a calibration set with 25\%, and a evaluation set with the last 25\%. Thereafter, each method is assessed on the evaluation set considering summary statistics of PIs width and coverage for $\alpha=0.1$ and a threshold explained variance of $\sigma_k=0.9$. All the data and code, which is written in Python can be found \href{https://github.com/Quilograma/ImprovedConformalizedQuantileRegression}{here}.

\begin{table}[H]
    \centering
    \begin{tabular}{cccc}
    \toprule
    Dataset & N & D & Source\\
    \midrule
    Blogfeedback & 60.000 & 280 & \cite{25}\\
         Boston house prices & 506 & 13 & \cite{26}\\
         Bike sharing & 17379 & 17 & \cite{27} \\
         \bottomrule
    \end{tabular}
    \caption{Datasets description.}
    \label{datasets}
\end{table}

\subsection*{Blogfeedback}
\label{blogfeedback}

\begin{figure}[H]%
    \centering
    \subfloat[\centering]{{\includegraphics[scale=0.2]{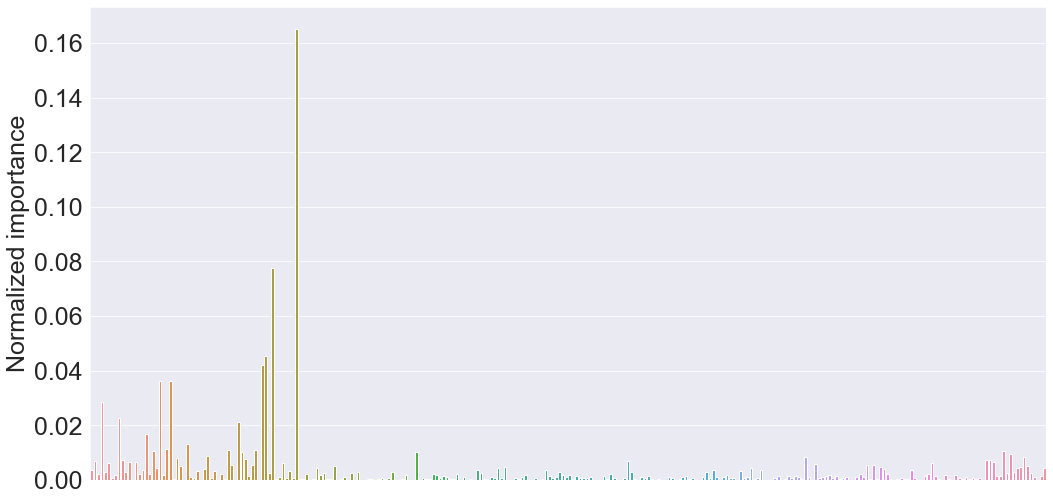} }}%
    \qquad
     \subfloat[\centering]{{\includegraphics[scale=0.2]{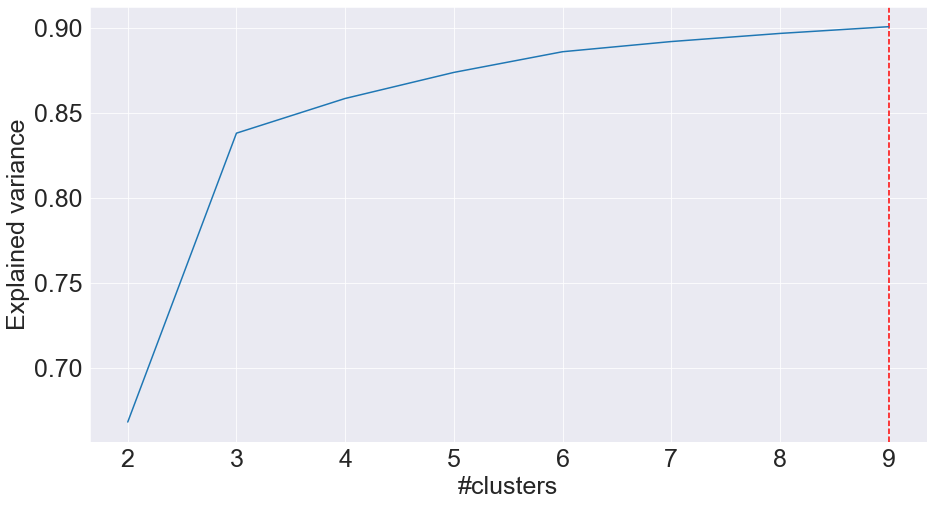} }}%
    \caption{(Blogfeedback) Feature importance (panel left). Number of clusters for a threshold fraction variance of 0.9 (panel right).}%
\end{figure}

\begin{table}[H]
    \centering
\begin{tabular}{ccccccccc}
\toprule
      Method & min &         max &      mean &       std &        Q1 &    median &        Q3 &       IQR \\
\midrule
Naive & 14.943970 &   20.984375 & 17.479034 &  0.976794 & 16.812687 & 17.417236 & 18.004509 &  1.191822 \\
 QR & 0.000002 & 1546.067383 &  5.822205 & 17.578970 &  1.105402 &  2.070409 &  4.677099 &  3.571697 \\
CQR & 0.054895 & 1552.663086 & 12.865065 & 17.589760 &  8.126973 &  9.508371 & 12.118934 &  3.991961 \\
 ICQR & 0.000061 & 1595.305754 & 23.701899 & 47.737405 &  2.431001 &  5.260328 & 35.063214 & 32.632213 \\
\bottomrule
\end{tabular}
\caption{(Blogfeedback) PI width summary statistics.}
\label{table2}
\end{table}
\begin{table}[H]
    \centering
\begin{tabular}{ccccccccc}
\toprule
     Method & min &      max &     mean &      std &       Q1 &   median &       Q3 &      IQR \\
\midrule
Naive & 0.899440 & 0.908237 & 0.903714 & 0.001837 & 0.902656 & 0.903872 & 0.904788 & 0.002132 \\
QR & 0.332600 & 0.828469 & 0.747061 & 0.092558 & 0.745585 & 0.778722 & 0.792366 & 0.046781 \\
CQR & 0.897908 & 0.906837 & 0.902400 & 0.002016 & 0.900756 & 0.902306 & 0.904105 & 0.003349 \\
ICQR & 0.894775 & 0.906371 & 0.901395 & 0.002302 & 0.899907 & 0.901706 & 0.902972 & 0.003065 \\
\bottomrule
\end{tabular}
\caption{(Blogfeedback) Coverage summary statistics.}
\label{table3}
\end{table}

\subsection*{Boston house prices}
\label{bostonhouseprices}

\begin{figure}[H]%
    \centering
    \subfloat[\centering]{{\includegraphics[scale=0.2]{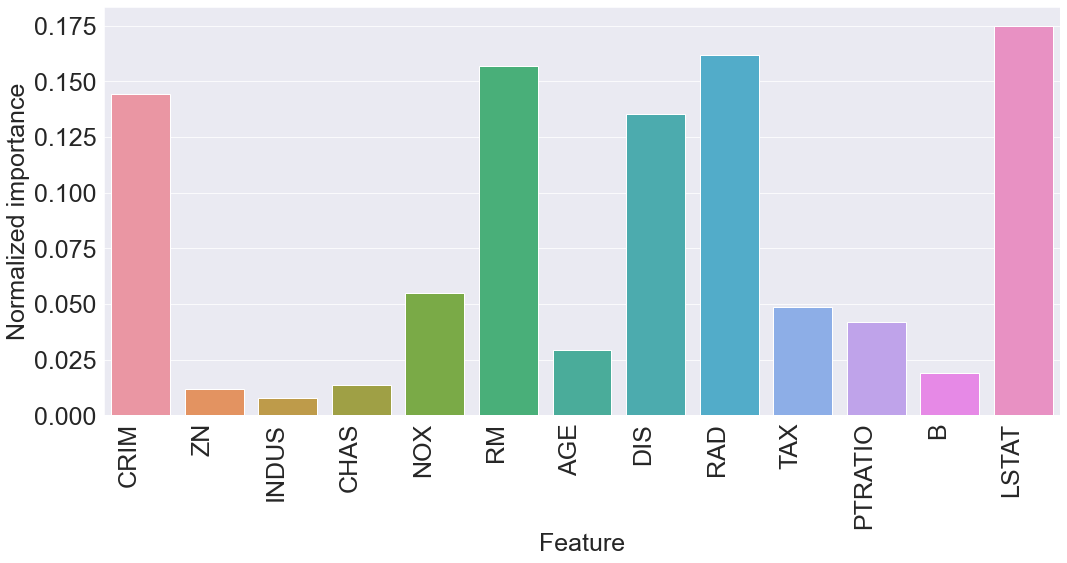} }}%
    \qquad
    \subfloat[\centering]{{\includegraphics[scale=0.22]{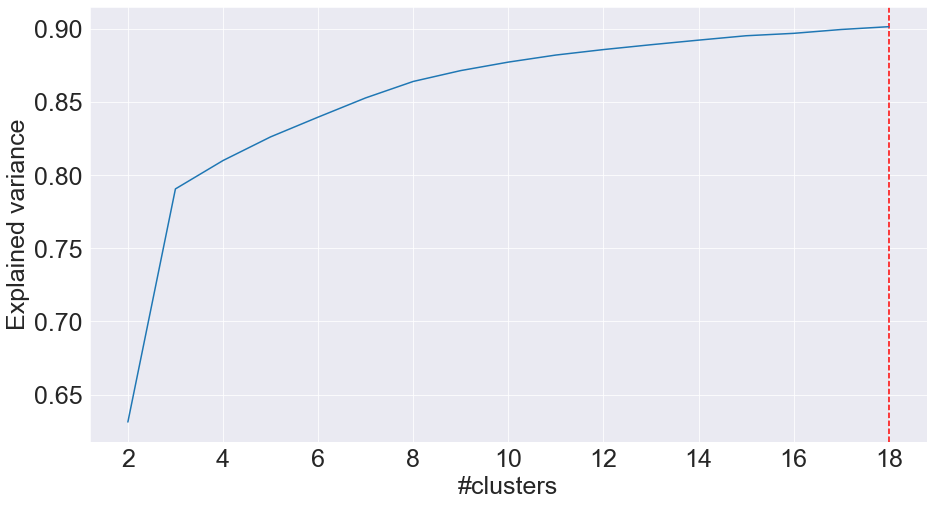} }}%
    \caption{(Boston house prices) Feature importance (panel left). Number of clusters for a threshold fraction variance of 0.9 (panel right).}%
\end{figure}

\begin{table}[H]
\centering
\begin{tabular}{ccccccccc}
\toprule
 Method & min &       max &      mean &       std &       Q1 &    median &        Q3 &      IQR \\
\midrule
Naive & 9.530071 & 12.689045 & 10.898064 & 0.602834 &10.478755
 & 10.807868
 & 11.260028 &
 0.781272 \\
QR & 1.324307 & 32.966770 &  8.323204 &  2.736615 & 6.490421 &  7.689121 &  9.650486 & 3.160065 \\
CQR & 2.978894 & 36.200993 & 10.921992 &  2.769641 & 9.070092 & 10.472181 & 12.346348 & 3.276256 \\
ICQR & 1.812310 & 44.346356 & 12.020735 &  4.937976 & 8.044880 & 10.293163 & 15.649653 & 7.604772 \\
\bottomrule
\end{tabular}
\caption{(Boston house prices) PI width summary statistics.}
\label{table4}
\end{table}

\begin{table}[H]
    \centering
    \begin{tabular}{ccccccccc}
\toprule
     Method & min &      max &     mean &      std &       Q1 &   median &       Q3 &       IQR \\
\midrule
Naive & 0.889764 & 0.976378 & 0.940472 & 0.018104 & 0.929134 & 0.944882 & 0.952756 & 0.023622 \\
QR & 0.771654 & 0.889764 & 0.837795 & 0.025148 & 0.818898 & 0.834646 & 0.858268 & 0.039370 \\
CQR & 0.866142 & 0.984252 & 0.938583 & 0.020290 & 0.929134 & 0.937008 & 0.952756 & 0.023622 \\
ICQR & 0.889764 & 0.984252 & 0.943228 & 0.019489 & 0.929134 & 0.944882 & 0.960630 & 0.031496 \\
\bottomrule
\end{tabular}
\caption{(Boston house prices) Coverage summary statistics.}
\label{table5}
\end{table}

\subsection*{Bike sharing}
\label{bikesharing}

\begin{figure}[H]%
    \centering
    \subfloat[\centering]{{\includegraphics[scale=0.2]{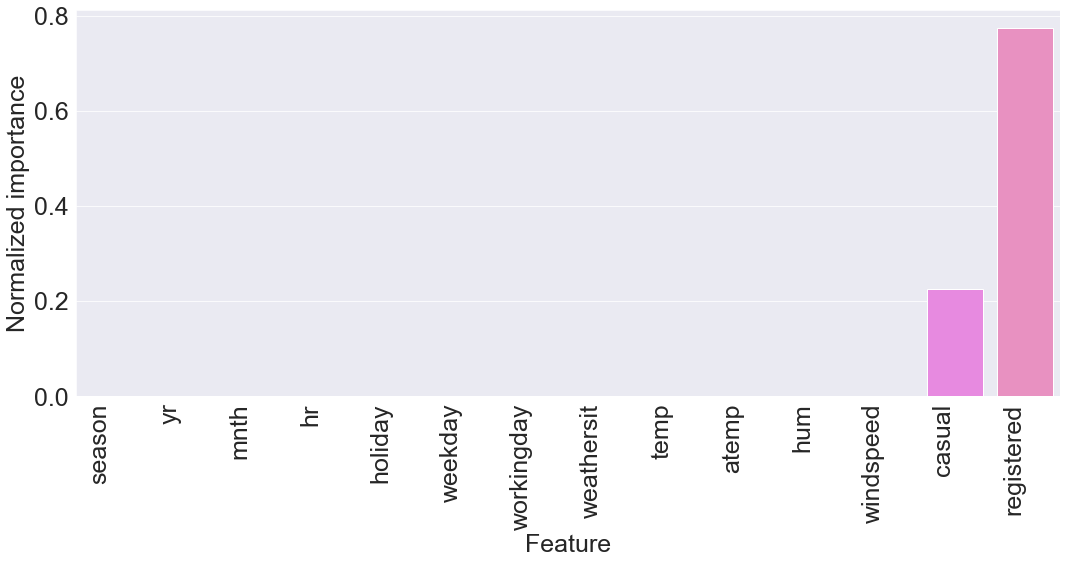} }}%
    \qquad
    \subfloat[\centering]{{\includegraphics[scale=0.22]{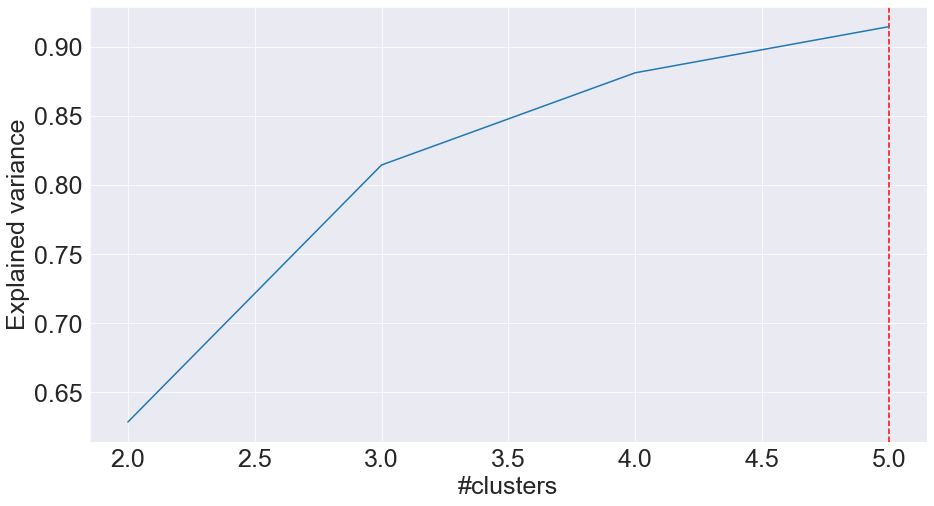} }}%
    \caption{(Bike sharing) Feature importance (panel left). Number of clusters for a threshold fraction variance of 0.9 (panel right).}%
\end{figure}

\begin{table}[H]
    \centering
    \begin{tabular}{ccccccccc}
\toprule
     Method & min &       max &     mean &      std &       Q1 &   median &       Q3 & IQR \\
\midrule
Naive & 0.427124 &  7.491333 & 2.336785 & 1.362940 & 1.327354 & 2.135223 & 3.189590 & 1.862236 \\
QR & 0.001831 & 35.566101 & 3.324998 & 2.115591 & 2.047175 & 2.682456 & 3.896203 & 1.849028 \\
CQR & 0.000229 & 36.159119 & 2.763314 & 2.186563 & 1.385670 & 2.202379 & 3.450058 & 2.064388 \\
ICQR & 0.000165 & 25.988984 & 2.249955 & 1.470346 & 1.267076 & 1.949806 & 2.848970 & 1.581894 \\
\bottomrule
\end{tabular}
\caption{(Bike sharing) PI width summary statistics.}
\label{table6}
\end{table}

\begin{table}[H]
    \centering
    \begin{tabular}{ccccccccc}
\toprule
     Method & min &      max &     mean &      std &       Q1 &   median &       Q3 & IQR \\
\midrule
Naive & 0.891139 & 0.915535 & 0.900700 & 0.004449 & 0.897583 & 0.900575 & 0.903625 & 0.006041 \\
QR & 0.084695 & 1.000000 & 0.897445 & 0.193131 & 0.887687 & 0.987457 & 0.995224 & 0.107537 \\
CQR & 0.887227 & 0.910702 & 0.900242 & 0.005158 & 0.897756 & 0.900230 & 0.904028 & 0.006272 \\
ICQR & 0.889068 & 0.916686 & 0.902656 & 0.006272 & 0.897986 & 0.902762 & 0.906847 & 0.008861 \\
\bottomrule
\end{tabular}
\caption{(Bike sharing) coverage summary statistics.}
\label{table7}
\end{table}
\subsection*{Results discussion}

In the two first datasets, QR is constantly undercovering; hence, QR bounds are in need of ICP to ensure $1-\alpha$ coverage. ICQR is clearly more adaptive to heteroscedasticity in comparison to CQR since it has generally higher std and IQR PI width, but lower median PI width, while still ensuring the same coverage, as seen in Table (\ref{table2}-\ref{table5}). 

In the last dataset (bike sharing), we have the opposite case, QR is overcovering as seen in Table (\ref{table7}). Consequently, we can reduce the bounds up to $1-\alpha$ coverage with ICP. 

\subsection*{Remarks}
\begin{itemize}
    \item Naive bounds are not dependent on the input in any form nor adaptive. The interval amplitude is always equal to $2\hat{q}$. The small deviation seen in the above tables is a result of training the model 100 times to reduce the variance associated with FFNN due to the random initialization process.
    
    \item Evaluating PIs by just looking to their mean value is not adequate. Most times, naive method has the lowest mean value; however, it achieves $1-\alpha$ coverage not in a group-balanced way, overcovering easy inputs, and PIs have always the same amplitude, ignoring heteroscedasticity. Therefore, analyzing PIs width by quantile gives a better perspective regarding conditional coverage. Adaptive methods generally have higher mean PI width because they are not optimized for heteroscedastic residuals. Nevertheless, the median PI width is usually lower in adaptive/heteroscedastic methods.     
\end{itemize}

\section{Conclusion}
\label{sec5}
In this paper, we have proposed an improved version of CQR. Results demonstrate that our version is further adaptive to heteroscedasticity, hence ICQR is one step ahead towards conditional coverage. The major shortcoming of ICQR in comparison to CQR are the two additional steps to calculate permutation importance and perform K-means. 
Despite this minor disadvantage, ICQR offers eye-catching adaptive PIs in comparison to the classic CQR, which convey us to strongly endorse its use across any high-stakes tabular regression problem.

\section*{Acknowledgments}
This work has been supported by COMPETE: POCI-01-0247-FEDER-039719 and FCT - Fundação
para a Ciência e Tecnologia within the Project Scope: UIDB/00127/2020.

\bibliographystyle{unsrt}  
\bibliography{references}

\end{document}